\documentclass[conference]{IEEEtran}
\IEEEoverridecommandlockouts

\usepackage{cite}
\usepackage{amsmath,amssymb,amsfonts}
\usepackage{algorithmic}
\usepackage{graphicx}
\usepackage{textcomp}
\usepackage{xcolor}
\usepackage[utf8]{inputenc}
\def\BibTeX{{\rm B\kern-.05em{\sc i\kern-.025em b}\kern-.08em
    T\kern-.1667em\lower.7ex\hbox{E}\kern-.125emX}}
\begin{document}

\title{IL-SLAM: Intelligent Line-assisted SLAM Based on Feature Awareness for Dynamic Environments\\
\thanks{This work was supported by the Asian Office of Aerospace Research and Development under Grant/Cooperative Agreement Award No. FA2386-22-1-4042, and  JST SPRING, Japan Grant Number JPMJSP2102.}
}

\author{
Haolan Zhang, Thanh Nguyen Canh, Chenghao Li, Ruidong Yang, Yonghoon Ji and Nak Young Chong
}

\maketitle

\begin{abstract}
Visual Simultaneous Localization and Mapping (SLAM) plays a crucial role in autonomous systems. Traditional SLAM methods, based on static environment assumptions, struggle to handle complex dynamic environments. Recent dynamic SLAM systems employ geometric constraints and deep learning to remove dynamic features, yet this creates a new challenge: insufficient remaining point features for subsequent SLAM processes. Existing solutions address this by continuously introducing additional line and plane features to supplement point features, achieving robust tracking and pose estimation. However, current methods continuously introduce additional features regardless of necessity, causing two problems: unnecessary computational overhead and potential performance degradation from accumulated low-quality additional features and noise. To address these issues, this paper proposes a feature-aware mechanism that evaluates whether current features are adequate to determine if line feature support should be activated. This decision mechanism enables the system to introduce line features only when necessary, significantly reducing computational complexity of additional features while minimizing the introduction of low-quality features and noise. In subsequent processing, the introduced line features assist in obtaining better initial camera poses through tracking, local mapping, and loop closure, but are excluded from global optimization to avoid potential negative impacts from low-quality additional features in long-term process. Extensive experiments on TUM datasets demonstrate substantial improvements in both ATE and RPE metrics compared to ORB-SLAM3 baseline and superior performance over other dynamic SLAM and multi-feature methods.
\end{abstract}


\section{Introduction}
SLAM plays a crucial role in autonomous mobile systems, enabling them to autonomously perceive and navigate in unknown environments. Due to the low cost of cameras, vision-based SLAM has gained favor in recent years. Traditional feature-based visual SLAM methods have dominated current visual SLAM approaches due to their high computational efficiency and robust performance in texture-rich static environments. However, traditional methods suffer from their static environment assumptions, making it difficult to correctly estimate poses in dynamic environments, leading to drift and failure.
To address this problem, approaches have evolved from initially using purely geometric methods to remove dynamic features as outliers~\cite{motion removal, point weighting, point correlations}, to recent hybrid approaches combining deep learning and geometric methods for dynamic point feature removal~\cite{DS-SLAM, DynaSLAM, Blitz-SLAM}. Through these methods, the accuracy and robustness of pose estimation in dynamic environments have been improved. However, this has introduced a noteworthy new problem: after dynamic feature removal in dynamic SLAM methods, the remaining available point features are not always sufficient for subsequent SLAM processing. To supplement the number of features remaining after dynamic removal, many recent methods have adopted the introduction of additional features (line features, plane features). 

In PLD-SLAM~\cite{PLD-SLAM}, the authors combine RGB images and depth images, obtaining semantic information through MobileNet to determine predefined dynamic regions, subsequently utilizing K-means~\cite{k-means} to cluster point and line features within predefined regions to remove potential dynamic point and line features as much as possible, then employing joint optimization of point and line features to improve camera pose estimation accuracy and robustness in dynamic scenes. DRG-SLAM~\cite{DRG-SLAM} also focuses on the problem of insufficient point features after removal, adopting the introduction of additional line and plane features for supplementation, making more comprehensive use of environmental geometric information. It then performs point sampling on lines and planes, employs SegNet~\cite{SegNet} to detect predefined dynamic objects, and combines epipolar constraints to remove features on predefined dynamic objects. For undefined dynamic objects, it removes point features through multi-view constraints, while the removal of lines and planes is determined by the number of sampling points that are dynamic points. Subsequently, during pose evaluation, it determines the conformity to the Manhattan World assumption to adopt either Manhattan World Pose Estimation or ordinary estimation methods. In PLDS-SLAM~\cite{PLDS-SLAM}, while supplementing point features, the authors also focus on dynamic removal, first applying geometric constraints to detected prior dynamic regions to obtain static features, while improving line segment matching through geometric constraints to enhance correct line matching. On the other hand, they employ Bayesian algorithms to track dynamic noise for continuous removal of dynamic noise from points and lines. YPL-SLAM~\cite{YPL-SLAM} detects potential dynamic regions, then utilizes epipolar constraints for further dynamic removal, subsequently extracting line features within the obtained static regions for subsequent pose estimation. The above methods adopt additional features to solve the problem of feature insufficiency caused by removal in dynamic environments. However, for the introduction of additional features, these methods always adopt a continuous introduction approach, that is, additional features are introduced regardless of the situation, even in static, low-dynamic, or dynamic situations where available features are sufficient. This brings two problems: (i) unnecessary computation, increasing the complexity of feature processing, and (ii) accumulation of low-quality additional features and noise during the introduction process, adversely affecting pose estimation and global optimization.

To address these two problems in existing multi-feature dynamic SLAM, we propose IL-SLAM, which employs a feature-aware mechanism to determine whether point features remaining after dynamic removal are sufficient, thereby determining the necessity of introducing additional features for supplementation while minimizing interference of additional features with the native point-based system. The key contributions are:
\begin{enumerate}
\item A feature-aware additional feature introduction mechanism that ensures minimal computational overhead by introducing line features only when point features are insufficient, avoiding unnecessary processing in scenarios where point features are adequate.
\item A minimal line feature processing strategy that participates in tracking and local mapping for pose correction but excludes line features from global optimization to prevent low-quality feature interference with long-term accuracy.
\item Extensive comparative experiments on TUM~\cite{TUM} datasets demonstrating significant improvements over state-of-the-art dynamic SLAM methods and trajectory error reduction compared to our baseline ORB-SLAM3.
\end{enumerate}

\section{METHODOLOGY}
\begin{figure*}[!htbp]
  \begin{center}
  \includegraphics[width=\linewidth]{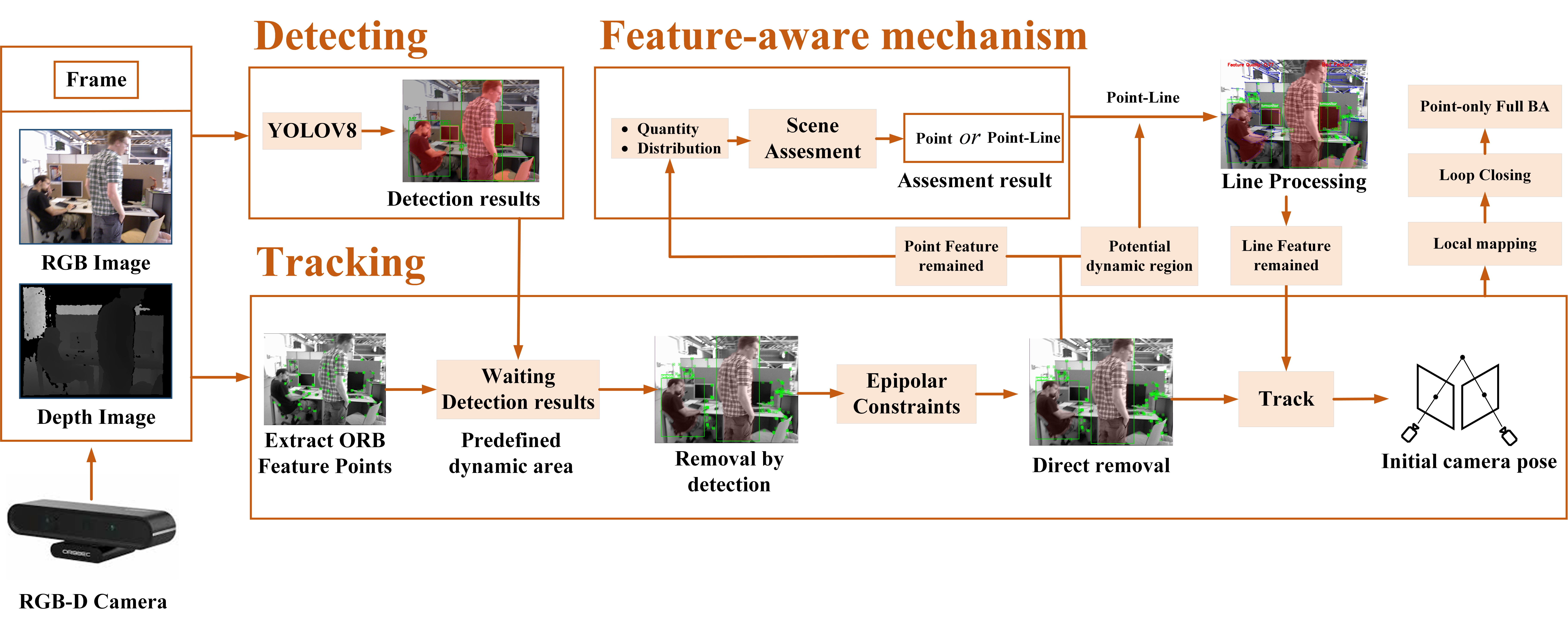}
  \caption{\textbf{Overview of our proposed IL-SLAM framework:} The framework is composed of three main units: feature-aware mechanism, Detecting and Tracking.}
  \label{fig:framework}
  \end{center}
\end{figure*}

The framework of our proposed method is shown in Fig.~\ref{fig:framework}. RGB-D images serve as input, simultaneously extracting ORB feature points and obtaining detection results using YOLOV8~\cite{Yolov8}. Subsequently, dynamic removal is performed on point features using predefined dynamic object regions detected and dynamic regions obtained through geometric constraints. The remaining point features are then fed into the feature-aware mechanism for evaluation of current point feature conditions, determining whether to activate point-line mode to introduce line features for supplementation. In point-line mode, current line features are extracted, dynamic removal is performed using the previously determined potential dynamic regions, and the static line features are then sent to tracking together with points for initial pose estimation. Finally, pose optimization is performed by points alone.
\subsection{Feature-Aware Mechanism}
For the design of the feature-aware mechanism, we are inspired by some static multi-feature methods' evaluation approaches for texture richness in the current environment~\cite{PLFG-SLAM, PLPF-VSLAM}. In dynamic situations, we evaluate available features from two aspects. We divide the current frame into $G$ (3×3) grids and evaluate the point feature situation after dynamic removal from two aspects: (i) the abundance of available feature points after removal, and (ii) the distribution of available features after removal. The calculation is as follows:
\begin{equation}
Q_{feature} = \frac{1}{G}\sum_{i=1}^{G}\left\{\frac{c_i}{c_{base}} + \frac{1}{1+\sqrt{\sigma_i^2}} \right\}
\label{eq:feature}
\end{equation}
where $G$ represents the number of grids, $c_{base}$ represents the base feature count, $c_i$ represents the feature count, and $\sigma_i^2$ is the variance of the feature distribution.

Subsequently, comparison with a threshold determines whether the current situation requires introduction of additional features:
\begin{equation}
\text{Scene} = \begin{cases}
\textbf{Point} & \text{if } Q_{feature} \geq th \\
\textbf{Point-Line} & \text{otherwise}
\end{cases} 
\label{eq:final_decision}
\end{equation}
where $th$ serves as the classification threshold. Both $c_{base}$ and $th$ are derived from relatively stable datasets (fr3/sitting/static in TUM and synchronous1,2 in BONN). We set $c_{base}$ and $th$ to the minimum value to ensure conservative classification.

Through the above proposed feature-aware mechanism, additional features can be introduced only when available point features are insufficient in dynamic situations, which controls the computational complexity of feature processing and reduces interference from low-quality additional features and noise.

\subsection{Line Extraction and Matching}
When the feature-aware mechanism determines that additional features are needed for supplementation (point-line mode), we employ LSD~\cite{LSD} to extract line features. For the extracted line features, following the representation method from~\cite{DRG-SLAM}, they are represented in 3D space as $L = (S_L, Q_L^1, M_L, Q_L^2, E_L)$, where these correspond to the start point, quarter point 1, midpoint, quarter point 2, and end point, respectively. Their projection on the image plane is denoted as $l = (s_l, q_l^1, m_l, q_l^2, e_l)$. 
For subsequent line segment matching, based on the above representation, we perform line feature matching with descriptor. Considering the auxiliary role of line features, the matching process adopts a simple matching approach. Given descriptors $(\ell_i, A_i, R_i)$ and $(\ell_j, A_j, R_j)$ representing length, angle, and edge strength response for line segments $l_i$ and $l_j$ respectively, the matching distance is computed by combining length difference, angle difference, edge strength difference. A line match is accepted when the distance is below a threshold and passes the ratio test.

In the local mapping stage, we employ our search-projection function to associate maplines with current frame observations using the same descriptor distance metric defined in line feature matching, where 3D maplines are projected into the image plane and matched within a spatial search window.

\subsection{Dynamic removal}
For dynamic removal, we adopt the most aggressive dynamic removal strategy to create worst-case scenarios for the point feature system, thereby validating the effectiveness of our system introduction while ensuring maximum removal of dynamic interference. Initially, we utilize previous detection results to perform the first step removal on all extracted point and line features. If a point feature $P_p$ falls within the predefined dynamic object region, it is directly removed. For line features $L$, the decision is made based on the above point sampling method: if three or more sampling points fall within the predefined dynamic object region, the entire line feature is completely removed, similar to the approach in~\cite{DRG-SLAM}.
\begin{figure}[!htbp]
  \begin{center}
  \includegraphics[width=0.9\linewidth]{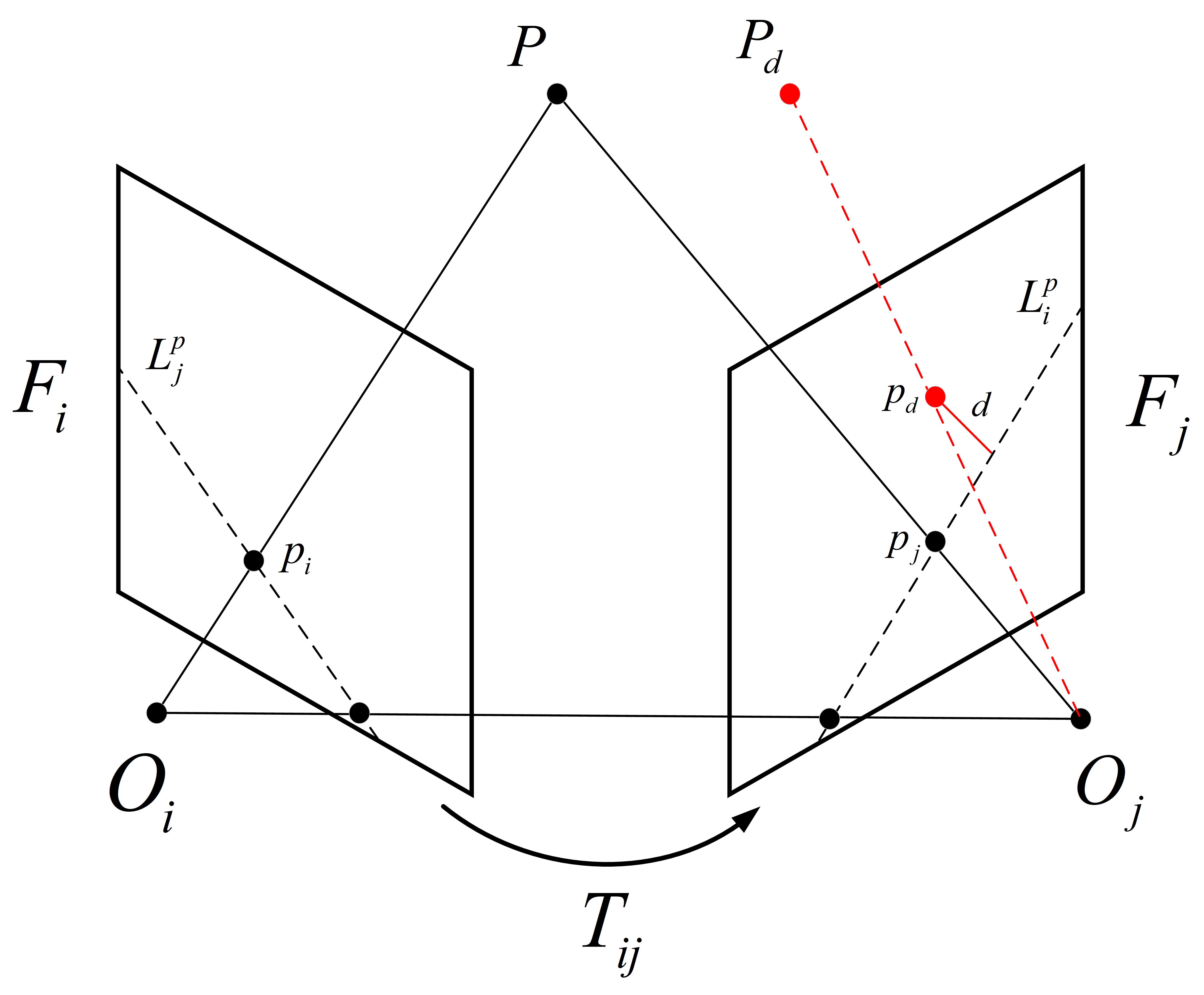}
  \caption{$O_i$ and $O_j$ denote the optical centers of the cameras, $P$ and $P_d$ are matched feature points in the images, including point features $P_p$ and line sampling points $S_L, Q_L^1, M_L, Q_L^2, E_L$. If $P$ is static, it should appear at positions $p_i$ and $p_j$, while $P_d$ represents the false position due to point motion. $T_{ij}$ represents the camera transformation between frames $F_i$ and $F_j$. The polar plane formed by $O_i$, $O_j$, and point $P$ intersects frames $F_i$ and $F_j$ at polar lines $L_i^p$ and $L_j^p$ (shown as dashed lines). The distance $d$ indicates the perpendicular distance from the feature point to its corresponding polar line.}
  \label{fig:epipolar}
  \end{center}
\end{figure}
After completing the first step removal based on detection results, we apply epipolar constraints to remove undefined dynamic features on the remaining features, as shown in Fig.~\ref{fig:epipolar}. For point features $P_p$, we directly compute their distance $d$ to the epipolar line and determine whether it exceeds the threshold. If it does, the point is directly removed. Following the work in~\cite{DS-SLAM}, we set the threshold to 1. For line features $L$, we adopt a similar approach by computing the distance to the epipolar line for each sampling point $S_L, Q_L^1, M_L, Q_L^2, E_L$ individually. Similar to the removal strategy in detection, if three or more sampling points exceed the distance threshold $d$, the entire line feature is completely removed. 
\label{fig:epipolar}

\subsection{Pose Estimation}

When the feature-aware mechanism determines that line features are needed, both point and line features participate in the tracking and local mapping processes through reprojection error minimization.

For point features, the reprojection error is computed as:
\begin{equation}
e_P^i = \|u_i - \pi(K \cdot T_{cw} \cdot P_i^w)\|^2
\end{equation}

For line features, the reprojection error is based on the distance between projected 3D line endpoints and observed 2D line endpoints:
\begin{equation}
e_L^i = \|s_l^i - \pi(K \cdot T_{cw} \cdot S_L^{w,i})\|^2 + \|e_l^i - \pi(K \cdot T_{cw} \cdot E_L^{w,i})\|^2
\end{equation}

where $u_i$ is the observed 2D point coordinates, $P_i^w$ is the corresponding 3D world point, $s_l^i$ and $e_l^i$ are the observed start and end points of line $i$, $S_L^{w,i}$ and $E_L^{w,i}$ are the corresponding 3D world coordinates of line endpoints, $T_{cw}$ is the camera pose, $K$ is the camera intrinsic matrix, and $\pi(\cdot)$ represents the projection function.

The pose estimation minimizes the combined reprojection error:
\begin{equation}
T_{cw}^* = \arg\min_{T_{cw}} \left( \sum_{i} \rho(e_P^i) + \sum_{i} \rho(e_L^i) \right)
\end{equation}

where $\rho(\cdot)$ is a robust kernel function. In the local optimization stage, both point and line features participate in bundle adjustment. However, following our hierarchical optimization strategy, global bundle adjustment uses only point features, as point features are generally sufficient at the global level, thereby preventing additional features from interfering with the native point-based system.

\section{EXPERIMENTAL RESULTS}
To comprehensively evaluate and validate our SRR-SLAM method, we conducted extensive experiments on TUM\cite{TUM} RGB-D datasets, encompassing diverse camera motion patterns including hemispherical, XYZ translation, RPY rotation, and static configurations across various indoor scenarios. We employ absolute trajectory error (ATE) and relative pose error (RPE) as primary evaluation metrics for trajectory accuracy assessment. The root mean square error (RMSE) and standard deviation (S.D.) are utilized to characterize both trajectory accuracy and system stability. The RPE evaluation, relative translation error (T.RPE) to provide comprehensive pose estimation analysis. All experiments were conducted on a computer equipped with Ubuntu 20.04 operating system, AMD Ryzen R7 CPU, NVIDIA RTX3080 GPU, and 32 GB of RAM.

\begin{figure*}[!htbp]
  \begin{center}
  \includegraphics[width=\linewidth]{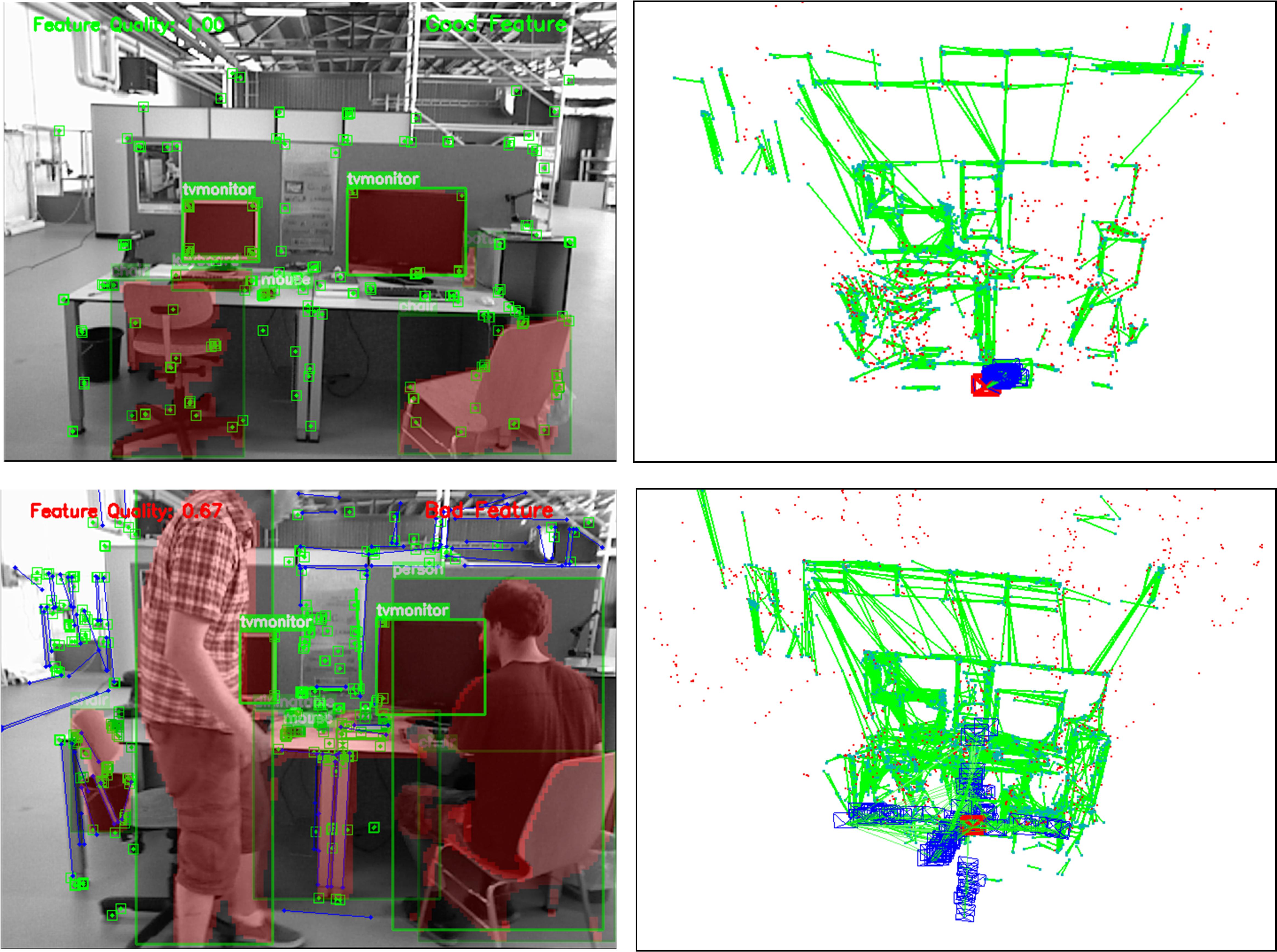}
  \caption{Performance visualization of our feature-aware mechanism on fr3/w/static (upper two images) and fr3/w/xyz (lower two images) sequences. The system adaptively switches between point mode in stable scenarios and point-line mode in complex dynamic environments based on feature quality assessment.}
  \label{fig:real_performance}
  \end{center}
\end{figure*}

Fig.~\ref{fig:real_performance} demonstrates the performance of our method on both a low dynamic sequence (fr3/w/static) and a complex dynamic sequence (fr3/w/xyz). From left to right, the upper two images and the lower two images represent the current frame visualization and mapping results for fr3/w/static and fr3/w/xyz, respectively. In the low-dynamic scenario (fr3/w/static), the first image shows the detection results with green feature points extracted from the current frame. The green text in the upper portion evaluates the current feature quality, providing a good feature assessment. The corresponding second image displays the sparse mapping results, retaining non-dynamic red map points and green map lines. The blue and red boxes represent the estimated camera poses. 

In contrast, the third and fourth images represent the complex dynamic scenario (fr3/w/xyz), where dynamic objects are correctly detected and features within human regions are removed. Due to the poor feature quality assessment after the dynamic removal (red feature quality with bad feature in the upper portion), our system introduces blue line features for supplementation, while the green line segments represent successfully matched line features in the current frame. The corresponding sparse mapping results show more map lines compared to fr3/w/static, as more line features were introduced. The red points represent map points, while the red and blue boxes indicate the estimated camera poses, consistent with the second images. This visualization clearly demonstrates how our feature-aware mechanism adapts to different environmental dynamics: utilizing point features efficiently in stable scenarios while intelligently supplementing with line features when needed in complex dynamic environments.

For validation, we comprehensively selected five representative sequences: fr3/w/half, fr3/w/rpy, fr3/w/xyz, fr3/s/xyz, and fr3/w/static. The first three sequences represent different camera motion patterns including translational, rotational, and combined translational-rotational movements, all accompanied by continuous motion of two persons in the environment. The fourth sequence features camera translation with two seated persons, while the last sequence involves a nearly stationary camera with minimal human motion in the environment. These sequences provide a comprehensive evaluation framework covering various combinations of camera dynamics and environmental motion complexity.

First, we compare our method with our baseline ORB-SLAM3~\cite{ORB-SLAM3} and several state-of-the-art dynamic SLAM methods from recent years: Blitz-SLAM~\cite{Blitz-SLAM}, RTD-SLAM~\cite{RTD} (referred to as RTD), and COEB-SLAM~\cite{COEB} (referred to as COEB). Blitz-SLAM employs semantic segmentation combined with epipolar geometry, RTD-SLAM integrates object detection with multi-view geometry, and COEB-SLAM combines semantic segmentation with optical flow information. The comparison results with these methods are presented in Table~\ref{tab:dynamic_compare}. Based on the overall results across the tested datasets, our method demonstrates outstanding performance, achieving the best results (highlighted in bold) in both ATE and T.RPE metrics across all five test sequences, consistently outperforming our baseline ORB-SLAM3 and other state-of-the-art dynamic SLAM methods.

In the three complex dynamic sequences (fr3/w/half, fr3/w/rpy, fr3/w/xyz), our method achieves the best performance with significant improvements over competing methods. For instance, in fr3/w/rpy, our ATE RMSE of 0.032 substantially outperforms ORB-SLAM3's 0.160, demonstrating the effectiveness of our aggressive dynamic removal strategy in eliminating dynamic interference. Remarkably, in scenarios with reduced dynamics (fr3/w/static, fr3/s/xyz), our method maintains optimal performance despite the aggressive removal strategy. In fr3/w/static, we achieve an ATE RMSE of 0.007, matching COEB-SLAM (which ranks second with underlined results) while significantly outperforming others. This demonstrates how our feature-aware mechanism enables flexible line feature supplementation to compensate for potential over-removal in low-dynamic scenarios.

The T.RPE results further validate the robustness of our approach, with our method consistently achieving the best translational accuracy across all sequences. For example, in fr3/w/half, our T.RPE of 0.016 outperforms all competing methods, while in fr3/w/xyz, we achieve 0.011 compared to ORB-SLAM3's 0.027. These results confirm that our feature-aware mechanism effectively maintains accurate pose estimation under varying environmental dynamics while minimizing computational complexity through selective feature introduction.
\begin{table*}[!htbp]
\centering
\caption{COMPARISON BETWEEN OUR IL-SLAM AND THE LATEST DYNAMIC SLAM SYSTEMS}
\label{tab:dynamic_compare}
\small
\vspace{-2mm}
\setlength{\tabcolsep}{10pt}
\begin{tabular}{l|cc|cc|cc|cc|cc}
\hline
\rule{0pt}{2.5ex} 
& \multicolumn{10}{c}{\textbf{ABSOLUTE TRAJECTORY ERROR (ATE/m)}} \\
\cline{2-11}
\rule{0pt}{2.5ex}
\hspace{-1mm}Sequences & \multicolumn{2}{c|}{ORB-SLAM3} & \multicolumn{2}{c|}{Blitz-SLAM} & \multicolumn{2}{c|}{RTD} & \multicolumn{2}{c|}{COEB} & \multicolumn{2}{c}{OURS} \\
& RMSE & S.D. & RMSE & S.D. & RMSE & S.D. & RMSE & S.D. & RMSE & S.D. \\
\hline
\rule{0pt}{2.5ex}
\hspace{-1mm}fr3/w/half & 0.231 & \textbf{0.009} & \textbf{0.025} & \underline{0.012} & \underline{0.028} & 0.024 & 0.028 & 0.014 & \textbf{0.025} & 0.014 \\
fr3/w/rpy & 0.160 & 0.073 & 0.035 & 0.022 & 0.167 & 0.030 & \underline{0.033} & \underline{0.020} & \textbf{0.032} & \textbf{0.019} \\
fr3/w/static & 0.024 & 0.012 & \underline{0.010} & 0.005 & 0.121 & \textbf{0.002} & \textbf{0.007} & \underline{0.003} & \textbf{0.007} & \underline{0.003} \\
fr3/w/xyz & 0.275 & 0.145 & \textbf{0.015} & \textbf{0.007} & 0.020 & 0.009 & \underline{0.016} & \underline{0.008} & \textbf{0.015} & \textbf{0.007} \\
fr3/s/xyz & \textbf{0.012} & \underline{0.006} & \underline{0.014} & \underline{0.006} & - & - & - & - & \textbf{0.012} & \textbf{0.005} \\
\hline
\rule{0pt}{2.5ex} 
& \multicolumn{10}{c}{\textbf{METRIC TRANSLATIONAL DRIFT (T.RPE/rad)}} \\
\cline{2-11}
\rule{0pt}{2.5ex}
\hspace{-1mm}fr3/w/half & \underline{0.024} & 0.016 & 0.025 & \underline{0.012} & 0.035 & 0.024 & 0.032 & 0.017 & \textbf{0.016} & \textbf{0.011} \\
fr3/w/rpy & 0.030 & 0.021 & 0.047 & 0.028 & \textbf{0.019} & \textbf{0.013} & 0.046 & 0.027 & \underline{0.026} & \underline{0.020} \\
fr3/w/static & 0.019 & 0.016 & 0.012 & 0.006 & 0.019 & 0.013 & \underline{0.009} & \textbf{0.003} & \textbf{0.006} & \textbf{0.003} \\
fr3/w/xyz & 0.027 & 0.020 & 0.019 & 0.009 & \underline{0.012} & \underline{0.007} & 0.021 & 0.011 & \textbf{0.011} & \textbf{0.006} \\
fr3/s/xyz & \textbf{0.009} & \underline{0.006} & 0.014 & 0.007 & - & - & - & - & \underline{0.010} & \textbf{0.005} \\
\hline
\end{tabular}
\vspace{0.2mm}
\begin{flushleft}
\hspace{5mm}\small \textbf{Note:} The best results of RMSE and S.D. are highlighted in bold, and the second best are underlined.
\end{flushleft}
\end{table*}

Additionally, we conducted comparisons with several multi-feature SLAM systems to demonstrate the advantages of our feature-aware approach over continuous multi-feature methods. O3L represents ORB-SLAM3 enhanced with line features, providing a direct comparison for evaluating the benefits of selective versus continuous line feature usage. Planar~\cite{Planar} represents a comprehensive SLAM system that integrates point, line, and plane features. DRG-SLAM~\cite{DRG-SLAM} combines point, line, and plane features, while YPL-SLAM~\cite{YPL-SLAM} integrates point and line features. Both DRG-SLAM and YPL-SLAM are capable of effectively handling dynamic interference, making them particularly relevant baselines for evaluating our approach in dynamic environments. The comparison results are shown in Table~\ref{tab:feature_compare}. Regarding ATE performance, the fr3/s/xyz sequence involves predominantly camera self-motion with minimal external dynamic interference. Under such stable conditions, our method does not continuously introduce additional features, resulting in a slight RMSE decrease compared to other SLAM systems that continuously incorporate additional features. However, our method achieves the best performance across the first four sequences containing dynamic object motion, with only a minor difference of 0.001 in standard deviation compared to the best result in the fr3/w/half sequence. These results demonstrate the effectiveness of our aggressive dynamic removal strategy combined with minimal introduction of additional features in reducing interference during complex scenarios. 

Regarding RPE performance, due to our aggressive dynamic removal strategy and selective introduction of additional features, our method's pose continuity performance is not optimal. However, compared to other multi-feature SLAM systems, our approach demonstrates results that are overall closest to O3L, which achieves the best performance in this metric. This proximity to the best-performing method validates that our selective feature introduction strategy effectively maintains pose continuity while minimizing interference from additional features, demonstrating the advantage of our feature-aware approach over continuous multi-feature methods.

\begin{table*}[!htbp]
\centering
\caption{COMPARISON BETWEEN OUR IL-SLAM AND THE EXISTING SLAM SYSTEMS USING PLANES OR LINES AS FEATURES}
\label{tab:feature_compare}
\small
\vspace{-2mm}
\setlength{\tabcolsep}{10pt}
\begin{tabular}{l|cc|cc|cc|cc|cc}
\hline
\rule{0pt}{2.5ex} 
& \multicolumn{10}{c}{\textbf{ABSOLUTE TRAJECTORY ERROR (ATE/m)}} \\
\cline{2-11}
\rule{0pt}{2.5ex}
\hspace{-1mm}Sequences & \multicolumn{2}{c|}{O3L} & \multicolumn{2}{c|}{Planar} & \multicolumn{2}{c|}{DRG-SLAM} & \multicolumn{2}{c|}{YPL-SLAM} & \multicolumn{2}{c}{OURS} \\
& RMSE & S.D. & RMSE & S.D. & RMSE & S.D. & RMSE & S.D. & RMSE & S.D. \\
\hline
\rule{0pt}{2.5ex}
\hspace{-1mm}fr3/w/half & 0.209 & 0.095 & 0.325 & - & \textbf{0.025} & - & 0.027 & \textbf{0.013} & \textbf{0.025} & \underline{0.014} \\
fr3/w/rpy & 0.158 & 0.077  & 0.553 & - & \underline{0.385} & - & 0.044 & \underline{0.025} & \textbf{0.032} & \textbf{0.019} \\
fr3/w/static & 0.020 & 0.011 & 0.293 & - & \textbf{0.007} & - & 0.009 & \underline{0.004} & \textbf{0.007} & \textbf{0.003} \\
fr3/w/xyz & 0.276 & 0.119 & 0.276 & - & \underline{0.018} & - & 0.026 & \underline{0.014} & \textbf{0.015} & \textbf{0.007} \\
fr3/s/xyz & \underline{0.009} & \textbf{0.005} & 0.024 & - & \textbf{0.008} & - & - & - & 0.012 & \textbf{0.005} \\
\hline
\rule{0pt}{2.5ex} 
& \multicolumn{10}{c}{\textbf{METRIC TRANSLATIONAL DRIFT (T.RPE/rad)}} \\
\cline{2-11}
\rule{0pt}{2.5ex}
\hspace{-1mm}fr3/w/half & \textbf{0.010} & \textbf{0.007} & 0.051 & - & \textbf{0.010} & - & 0.023 & 0.012 & \underline{0.016} & \underline{0.011}\\
fr3/w/rpy & \textbf{0.010} & \textbf{0.006} & 0.051 & - & 0.042 & - & 0.044 & 0.030 & \underline{0.026} & \underline{0.020} \\
fr3/w/static & \textbf{0.004} & \textbf{0.003} & 0.023 & - & 0.004 & - & 0.009 & 0.004 & \underline{0.006} & \textbf{0.003} \\
fr3/w/xyz & \textbf{0.009} & \textbf{0.006} & 0.036 & - & 0.009 & - & 0.019 & \underline{0.009} & \underline{0.011} & \textbf{0.006} \\
fr3/s/xyz & \textbf{0.006} & \textbf{0.003} & 0.009 & - & \underline{0.007} & - & - & - & 0.010 & \underline{0.005} \\
\hline
\end{tabular}
\vspace{0.2mm}
\begin{flushleft}
\hspace{6mm}\small \textbf{Note:} The best results of RMSE and S.D. are highlighted in bold, and the second best are underlined.
\end{flushleft}
\end{table*}

\begin{figure*}[!htbp]
  \begin{center}
  \includegraphics[width=\linewidth]{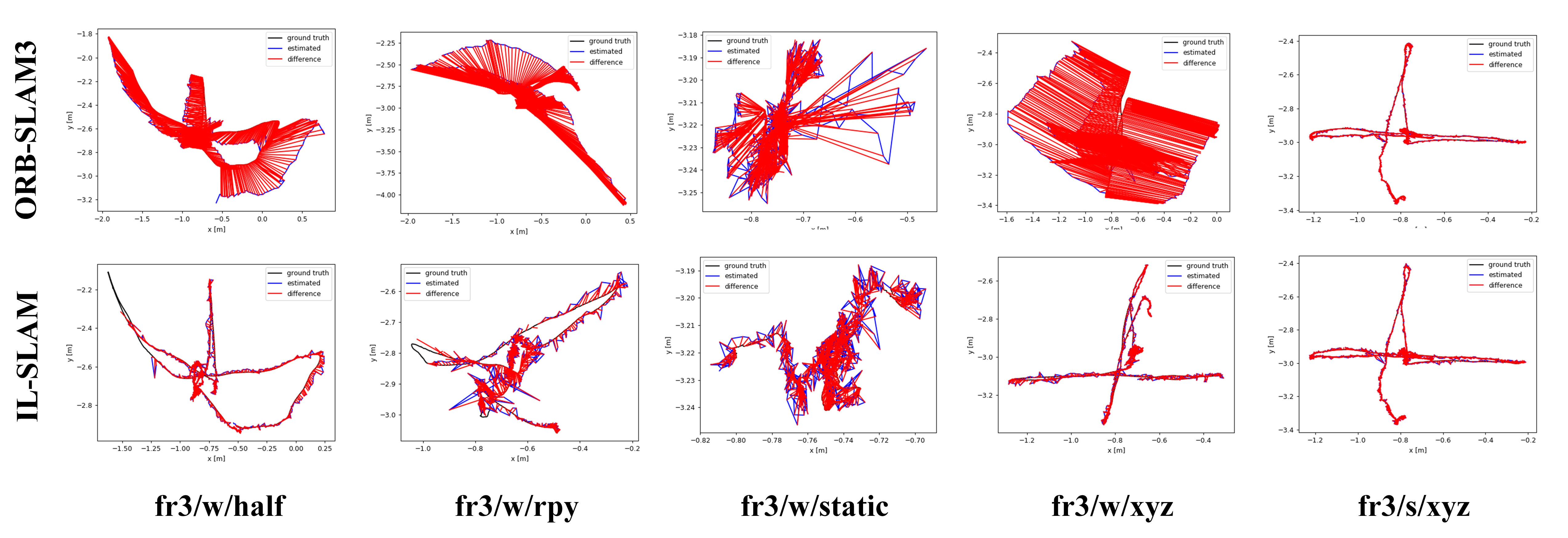}
  \caption{ATE results of IL-SLAM and ORB-SLAM3 running in five sequences, where the black line represents the true trajectory, the blue line represents the trajectory estimated by the algorithm, and the red line represents the difference between the estimated and true values}
  \label{fig:Trajectory_compare}
  \end{center}
\end{figure*}

\begin{figure*}[!htbp]
  \begin{center}
  \includegraphics[width=\linewidth]{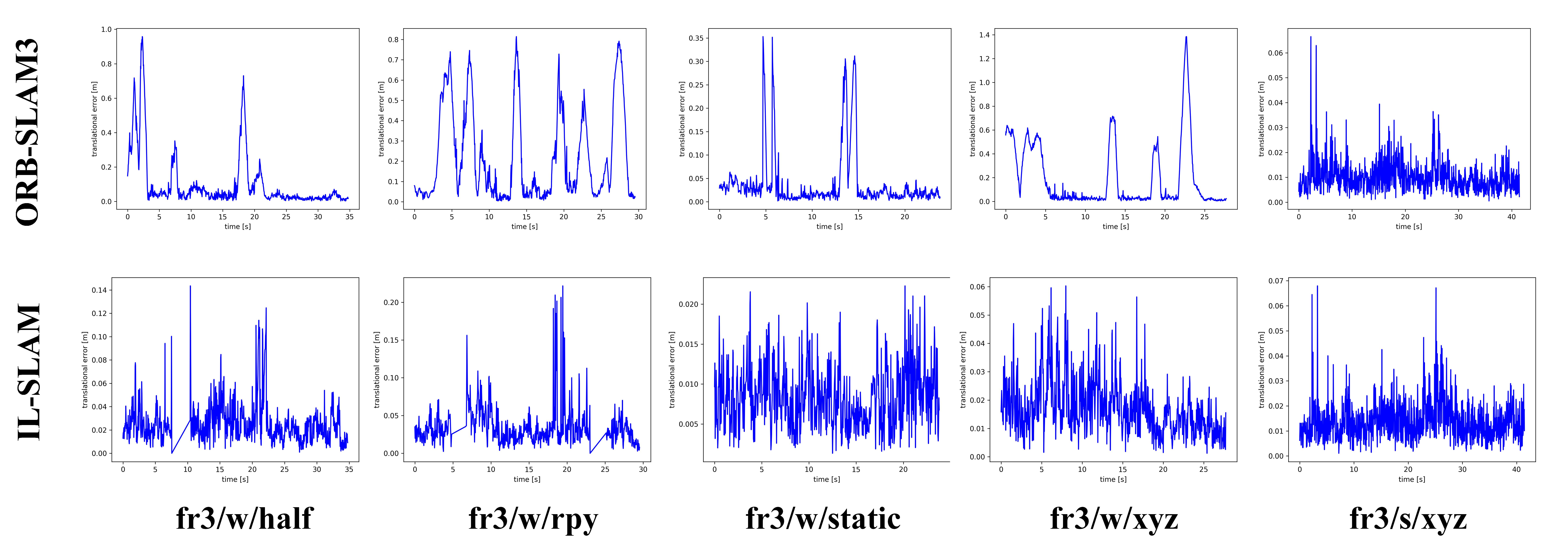}
  \caption{RPE results of IL-SLAM and ORB-SLAM3 running in five sequences, where the blue lines represent the RPE results for each time point.}
  \label{fig:Traj_ape}
  \end{center}
\end{figure*}

Fig.~\ref{fig:Trajectory_compare} and Fig.~\ref{fig:Traj_ape} present trajectory comparisons and RPE error analysis between our method and baseline ORB-SLAM3 across the five evaluated sequences. The results reveal that ORB-SLAM3 suffers from complete estimation failure in the first four dynamic sequences due to dynamic interference, with maximum RPE errors reaching 0.35-1.4. In contrast, our IL-SLAM method achieves robust pose estimation across these four sequences, maintaining RPE errors within the range of 0.06-0.2, demonstrating the effectiveness of our dynamic interference handling approach. Notably, in the final sequence with minimal dynamic interference, the trajectories of ORB-SLAM3 and IL-SLAM nearly perfectly overlap, with both methods achieving RPE errors around 0.06. This convergence in stable conditions validates that our method's minimal processing of additional features effectively prevents interference while preserving the native system's inherent accuracy when environmental dynamics are absent.

\section{CONCLUSIONS}

This paper presents IL-SLAM, a feature-aware line-assisted SLAM system that introduces a feature-aware mechanism that intelligently evaluates point feature sufficiency after dynamic removal and selectively activates line features only when necessary. This approach effectively resolves two critical issues in current multi-feature dynamic SLAM systems: unnecessary computational overhead from continuous feature introduction and potential performance degradation from low-quality additional features in scenarios where point features are already sufficient.

Extensive experiments on TUM RGB-D datasets demonstrate that our method consistently outperforms baseline ORB-SLAM3 and state-of-the-art dynamic SLAM methods, achieving the best ATE performance in four out of five test sequences. The hierarchical optimization strategy, where line features participate in local mapping but are excluded from global bundle adjustment, successfully balances tracking robustness with long-term accuracy. The experimental results validate that selective feature introduction based on actual need outperforms continuous multi-feature approaches, particularly in complex dynamic environments, while maintaining computational efficiency through minimal additional feature processing.

While our current binary decision mechanism has proven effective, future work will explore more sophisticated evaluation strategies. Rather than the threshold-based approach, we plan to consider more comprehensive environmental factors following approaches like QualiSLAM~\cite{QualiSLAM}. This would enable jointly evaluating point features and additional features from a unified quality perspective, considering their individual and collective contributions to system performance. Such an approach could achieve better balance between computational efficiency and tracking accuracy through more informed feature utilization decisions based on broader contextual information.

\end{document}